# Solving the Class Imbalance Problem
# Using a Counterfactual Method for Data Augmentation

**Mohammed Temraz[1,2] & Mark T. Keane[1,2,3]**

School of Computer Science, University College Dublin, Belfield, Dublin 4, Ireland

Insight Centre for Data Analytics, University College Dublin, Belfield, Dublin 4, Ireland

VistaMilk SFI Research Centre, University College Dublin, Belfield, Dublin 4, Ireland

mohammed.temraz@ucdconnect.ie

mark.keane@ucd.ie

## Abstract

Learning from class imbalanced datasets poses challenges for many machine learning algorithms. Many real-world domains are, by definition, class imbalanced by virtue of having a majority class that naturally has many more instances than its minority class (e.g., genuine bank transactions occur much more often than fraudulent ones). Many methods have been proposed to solve the class imbalance problem, among the most popular being oversampling techniques (such as SMOTE). These methods generate synthetic instances in the minority class, to balance the dataset, performing data augmentations that improve the performance of predictive machine learning (ML) models. In this paper, we advance a novel data augmentation method (adapted from eXplainable AI), that generates synthetic, counterfactual instances in the minority class. Unlike other oversampling techniques, this method adaptively combines existing instances from the dataset, using actual feature-values rather than interpolating values between instances. Several experiments using four different classifiers and 25 datasets are reported, which show that this Counterfactual Augmentation method (CFA) generates useful synthetic datapoints in the minority class. The experiments also show that CFA is competitive with many other oversampling methods, many of which are variants of SMOTE. The basis for CFA's performance is discussed, along with the conditions under which it is likely to perform better or worse in future tests.

# 1   Introduction

Imbalanced datasets create significant problems for machine learning (ML) in classification tasks [47]. Classically, this problem arises in binary classification tasks when most of data comes from one class (i.e., the *majority class*) and less comes from the other class (i.e., the *minority class*). For instance, in credit-card fraud-detection, datasets always have many more non-fraudulent instances than fraudulent ones, simply because the latter are rarer than the former [11]. When a given class is under-represented in the dataset in this way, a classifier's performance can be compromised in several ways; for instance, it may show poor accuracy in predicting the minority class, or spuriously high accuracy for the classifier as a whole (based only on its success with the majority class), and/or it can result in poor rule induction for decision trees [11, 25, 32, 40, 57]. This *class-imbalance problem* has been recognized in many real-world application domains such as medical diagnosis [66], fraud detection [11], text classification [78], and detection of oil spills in satellite radar images [42]. Notably, recently, some of the techniques proposed to solve the class imbalance problem have also proved useful in data augmentation for deep learning models, when new, synthetic data-points need to be generated to create the large, labelled datasets required for better performance [63, 73, 74].

In this literature, several important approaches have emerged to deal with this problem based on solutions at the data or algorithm levels. *Data level* solutions attempt to change the distribution of the imbalanced data by re-sampling the original data [9, 10, 11, 12, 27, 30, 31, 46, 58, 59]; typically, these techniques either oversample the minority class or undersample the majority class or sample using some combination of both of these methods. Specifically, the *Synthetic Minority Over-Sampling Technique* (SMOTE) has become a very popular method for solving class-imbalance issues in traditional ML and has also been applied to the data augmentation problem in deep learners [11, 31, 46, 63]. *Algorithm level* solutions aim to modify the machine learning algorithms used, to mitigate their bias towards majority groups; typically, these techniques involve the use of cost-sensitive and ensemble methods [20, 24].

In this paper, we explore a novel approach to both the class imbalance and data augmentation problems using an *instance-based counterfactual method* that generates synthetic data-points in the minority class [39, 64]; interestingly, this method was previously developed to solve problems in eXplainable AI (XAI; for reviews see [37, 38]). In logic, Lewis [45] proposed that counterfactuals are the *closest possible world* to the current world in which the outcome is different. Hence, the intuition behind the current technique is that it generates "synthetic" counterfactual instances using the actual feature-values of instances (not interpolated values) that are the close to existing instances thus populating the minority class with plausible adaptations of existing data. If this intuition is correct then the synthetic instances generated by this counterfactual method should improve ML performance, perhaps to a level that advances current class-imbalance techniques.

## 2   Related Work

The related work to the present research comes from two different strands of AI research: from (i) data-level sampling techniques for the class imbalance problem and (ii) counterfactual methods for eXplainable AI (XAI).

Data level solutions to the class imbalance problem are dominated by three main approaches: Random Over-Sampling (ROS), Random Under-Sampling (RUS), and Synthetic Minority Over-Sampling Technique (SMOTE) [11, 31, 46]. In ROS, the class distribution is balanced by randomly adding multiple copies of some of the minority classes to the training data [46]. Whereas, with RUS, a certain number of examples of the majority class are randomly removed from the original dataset [31]. Although these methods can re-balance the original dataset, they have some drawbacks. Since ROS merely copies minority-class instances, no new information is added to the dataset and, hence, it can lead to overfitting [72]. On the other hand, since RUS randomly removes examples from the majority class, data can be discarded that may be important [53]. The third option – SMOTE -- adopts a somewhat different approach based on oversampling from the minority class. As SMOTE is the baseline method used for comparisons in the present experiments, we briefly describe it in more detail here (see section 2.1) along with important SMOTE variants (see section 2.2) before going on to describe the counterfactual method we have adapted from the XAI literature (see section 2.3). Finally, we briefly sketch the recent and very small literature that has begun to apply these counterfactual XAI methods to class-imbalance and data augmentation problems (see section 2.4).

## 2.1 Data Sampling Methods for the Class Imbalance Problem: SMOTE

*Synthetic Minority Oversampling Technique* (SMOTE) oversamples the minority class by creating "synthetic" instances rather than by oversampling using replacement. It is one of the most widely used solutions to the class-imbalance problem; Google Scholar lists over 14,500 citations to the original paper [11, 21]. In SMOTE, the new example in the minority class is created by interpolating between several minority class instances. By interpolating instead of copying instances, SMOTE avoids the over-fitting problem, and creates new synthetic instances in neighborhoods surrounding instances in the minority class. Briefly, the algorithm works as follows. Assume that the minority class is $P$ and the majority class is $X$. SMOTE starts by randomly selecting a minority instance, $p_i$, from the minority class, $P$, and then determines $m$, as the nearest neighbors of $p_i$. After determining $m$ nearest neighbors of $p_i$, it selects a random neighbor $m'$: where $m' \in P$. Finally, SMOTE creates a new instance $p_{new}$ using the following formula:

$$p_{new} = p_i + (m' - p_i) \times \delta, \delta \in [0,1] \qquad (1)$$

where $\delta$ is a random number between 0 and 1. This new instance is then added to the dataset for the minority class. One of the potential problems with SMOTE is that its generation of minority instances is done without reference to the majority class or, indeed, any consideration that some minority instances may be better than others to use in this data-generation process. Another issue is that it may introduce noise, by generating interpolated values that do not exist in the domain (e.g., the interpolated value could be out-of-distribution). Accordingly, many extensions have been made to SMOTE that improve on its operation. In the following sub-section, we review the SMOTE variants that are closest to the current method proposed, to reveal how it differs.

## 2.2 SMOTE Variants: Three Key Insights

There are many variants of SMOTE that improve on the original's performance based on several insights about how to solve the class-imbalance problem; so, these variants often hinge on identifying important/safe regions in the minority class, they emphasise the importance of focusing on the border region between the majority and minority classes, and sometimes analyze the majority class with respect to the minority to guide under/oversampling.

**SMOTE in Selected Regions.** One critical improvement to the original SMOTE method hinges on the insight that not all regions in the minority class are equal, some may be more important or safer than others within which to apply SMOTE. For instance, k-Means SMOTE [19] clusters minority instances into $k$ clusters and then oversamples from clusters with the most instances, the assumption being that these are safer regions and are less likely to generate noise (see AND-SMOTE [76] for a related solution). Other versions of this approach have used DBSCAN, a based clustering algorithm, to identify safe regions [10] or use representative points within clusters to guide SMOTE (e.g., CURE-SMOTE [49]). Some methods project the minority class into a lower dimension before applying SMOTE to the clusters found; for instance, SOMO [18] uses a self-organizing map to transform high-dimensional datasets into a two-dimensional space, and LLE-SMOTE [71] uses a locally-linear embedding algorithm to project into a lower dimension where the datasets are more separable. Still others, such as G-SMOTE and ADASYN, explore different ways to identify regions within which to generate minority instances. G-SMOTE [61] defines a geometric region around each minority class instance for generating synthetic datapoints. ADASYN, proposed by He et al. [30], generates minority class instances according to their distributions, generating more synthetic data from minority instances that are *harder to learn* compared to minority instances that are *easier to learn* (where ease-of-learning is related to the number of instances in the *k*-nearest neighbors that belong to the majority class). However, these safe-region solutions owe a lot to another key insight, namely that regions close to the class boundary are particularly important for instance generation.

**SMOTE On the Borderline.** The idea that different regions in the dataset need to be handled differently, owes a lot to the intuition that minority instances close to the decision boundary of the classifier are particularly important to successful classification, so generating minority instances in this boundary region should help performance more. Borderline-SMOTE (B-SMOTE), proposed by Han et al. [27], realized this idea by creating instances using only minority instances that are close to the decision boundary. So, again, assume that the minority class is *P* and the majority class is *X*, and the whole training set is *T*. In B-SMOTE, for every minority instance $p_i$ in the minority class, *P*, the method calculates its $m$ nearest neighbors from the training set *T*. It should be noted that the number of majority instances among the $m$ nearest neighbors is called as $m'$ ($0 \leq m' \leq m$). In step 2, if all the $m$ nearest neighbors of $p_i$ are majority instances (i.e., $m = m'$), $p_i$ is considered as noise and is not used in the next step. If the set of $p_i$'s majority nearest neighbors is larger than that of its minority ones (i.e., $m/2 \leq m' < m$), $p_i$ will be easily misclassified and put into a DANGER set. If $0 \leq m' < m/2$, then $p_i$ is safe and does not participate in the subsequent steps. This DANGER set contains the borderline instances of the minority class *P*. Finally, for each instance in the DANGER set, the *k*-nearest neighbors from *P* are found and the steps from the original SMOTE method are applied to them to generate synthetic instances in the minority class. This insight about the importance of the boundary regions has been exploited in different ways. For example, SVM-SMOTE [54] uses an SVM to approximate the decision boundary and then generates new synthetic data along the lines joining each minority-class support-vector with its nearest neighbors using interpolation or extrapolation techniques. In a similar vein, M-SMOTE [34] divides minority instances into three groups (security instances, border instances, and latent noise instances) and then treats these groups differently when generating instances. Other variants in this vein, adjust the sampling rate for some minority instances (e.g., those close to the boundary) to improve these methods further (see GAS-SMOTE [36] and SMOTE-D [68]). This use of boundary regions in the minority class also raises questions about the relationship of majority instances to minority instances leading to a third insight underlying SMOTE variants; namely, that oversampling in the minority class can be informed/guided by considering the majority class.

**Using the Majority Class.** A third important insight in this research area, which becomes more apparent when borderlines are explored, is the idea that the relationship between the majority class and the minority class can also help guide SMOTE. Earlier, we saw that Borderline-SMOTE does not interpolate instances when the $k$-nearest neighbors show a preponderance of majority instances (see also ADYSYN). This is one way to take the majority class into account. Other methods explore the relationship between classes to undersample the majority class using data-cleaning techniques [5], or to guide the oversampling of the minority class [6, 9]. For instance, SMOTE-Tomek [5] finds pairs of instances between the minority and majority classes that are very similar (low Euclidean distance) – a Tomek Link – and then removes the majority instance in the pair; by removing these Tomek-links and applying SMOTE to the minority class, it attempts to re-balance the classes (SMOTE-ENN [5] uses a related data-cleaning approach involving the Edited Nearest Neighbour method). Other methods, SL-SMOTE [9] and SWIM [6] perform explicit analyses of the majority class and use this analysis to inform/guide minority instance generation. Safe-Level-SMOTE (SL-SMOTE) [9] does this by computing a *safe-level score* for each minority instance (where safety is based on the frequency of majority instances in the $k$-nearest neighbours) and a *safe-level ratio* based on the safe-level score of a minority instance over that of its neighbours. SL-SMOTE's finer analysis of the relationship between majority and minority has been shown to improve performance over B-SMOTE. Sampling WIth the Majority (SWIM) [6] adopt a different approach, leveraging information about the density of well-represented majority instances (using the Mahalanobis distances), requiring generated minority instances to have similar distances to their minority seeds. So, SWIM essentially analyses the topology of the majority class to guide the generation of minority instances (see MC-RBO [41] and LN-SMOTE [50] for related approaches). Finally, SMOTE-RSB [59] is another method that takes similarities to majority instances into account, in computing Rough Sets over the minority class, after SMOTE has been applied to generate additional minority instances; this method acts like a data-cleaning step to remove generated instances that might be noise. Many of these methods improve on B-SMOTE's performance and, as such, show that paying more attention to the majority class can play a key role in informing/guiding instance generation in the minority class. We will see later that while the current counterfactual methods reflect these three key insights about out to improve on SMOTE it is quite different from all of the above methods in how it operates, (see section 3.3). But, before considering this counterfactual method in detail, we first briefly review how it has emerged in XAI.

## 2.3 Counterfactual Generation in XAI

In this paper, we deploy a case-based counterfactual method to generate synthetic, minority-class instances [39, 64]. Counterfactual methods have been developed to generate post-hoc examples to explain the predictions of black-box ML models and to provide algorithmic recourse for end-users trying to mitigate automated decisions (for reviews see [37, 38]). The classic counterfactual explanation is one that is given when an automated system refuses a person on a loan application [70]; when the end-user asks "why?" the system might counterfactually explain that "If you requested a loan for $500 less over a shorter term, then you would have been granted the loan". That is, the counterfactual explanation tells users about the conditions under which outcome would change, the closest world to their world in which the outcome would be what they desire.

Counterfactuals have been researched for some time in AI under diverse names; for instance, in the past, they have been called Nearest Unlike Neighbours (NUNs [16, 51, 55]) or inverse classifications [1, 43]. Recently, they have emerged as a hot topic in XAI because they appear to have psychological benefits (i.e., people naturally understand them) and legal benefits (i.e., they are said to be GDPR compliant). Optimization techniques are currently the most popular method for computing counterfactuals [14, 52, 62, 70]. Given a test instance (e.g., one encoding the original loan refusal) these optimization methods search a (sometimes randomly) generated space of perturbations of the query (i.e., synthetic instances) under a loss function that balances proximity to the test-instance against proximity to the decision boundary for the counterfactual class (i.e., the class that counters that of the query), using a scaled $L_1$-norm distance-metric. Wachter, et al.'s [70] seminal method uses gradient descent to find the best counterfactual instance for a query, though later models have used other techniques (e.g., genetic algorithms). Mothilal, et al. [52] proposed the *Diverse Counterfactual Explanations* (DiCE) method as an extension, to generate a set of *diverse* counterfactual candidates avoiding the problem of generating sets of candidates that were trivial variants on one another. The main problem with these optimization methods is that, given their "blind" perturbation of test-items, they sometimes generate out-of-distribution, invalid data-points [17, 44, 70]. This defect has potentially serious side-effects for their use in the class-imbalance problem, as it suggests that they might populate the minority class with noise, with consequential negative effects on a classifier's performance.

However, a very different cases-based approach to counterfactual generation has recently been proposed [39, 64]. This *instance-guided method* finds the test-instance's nearest-neighbor that takes part in a so-called *explanation case (xc)*. An explanation case captures a counterfactual relation between existing instances in the dataset, that are in opposing classes either side of a decision boundary, with the constraint that the pair of instances differ by at most two feature-differences. For example, the loan dataset could contain two existing cases that are counterfactually related; one about a "30-year old, female accountant earning *$35k* who was *refused* a $10k loan" that is counterfactually related to another instance with a different outcome, namely a "30-year old, female accountant earning *$40k* who was *granted* a $10k loan" (differences shown in italics). This explanatory case implicitly suggests that "IF one earns *$40k* rather than *$35k* THEN the loan decision is likely to be *granted* rather than *refused*". So, if I am a "30-year old, male teacher earning $35k who was refused a $10k loan" then this algorithm could find this explanatory-pair as a nearest neighbour and suggest that if this male-teacher earned $5k more ($40k rather than $35k) then the $10k loan would be granted". In the XAI context, this method has been shown to generate close, plausible counterfactuals and appears to avoid the out-of-distribution pitfalls that arise in optimization techniques (see [17, 39, 64]). From a data augmentation perspective, this method can be seen as supporting the creation of synthetic data-points in the minority class, using information from these known counterfactual pairs. However, few XAI techniques have been applied to data augmentation problems. In the next subsection, we briefly sketch this small literature on the topic.

## 2.4 Using Counterfactuals for Data Augmentation

Beyond XAI, our hypothesis is that counterfactual methods can also play a role in data augmentation to solve class-imbalance problems, that generated synthetic counterfactual cases could improve the predictive accuracy of AI models. Although, there are now 100s of papers on counterfactuals in XAI, only a handful of papers consider their use in data augmentation [29, 56, 65, 77]. In evaluating XAI counterfactual methods, Mothilal et al. [52] suggested that a good method should generate a set of counterfactuals that can substitute for the original dataset (calling it *substitutability*); that is, if the set of generated counterfactuals were plausible and close to the original data then their predictive performance should parallel that of the original dataset. However, Mothilal et al. did not consider using their counterfactual method for data augmentation. In a student project, Hasan [29] did and tried to determine whether an augmented dataset based on generated counterfactuals could act as a proxy dataset, but only found modest success.

A selection of other papers in diverse areas have also circled the issue of using counterfactual techniques for data augmentation. Subbaswamy and Saria [65] considered the problem of *dataset shift*, where there is a divergence between the context in which a model was trained and tested; they use the notion of "counterfactual risk" to diagnose this problem using causal models. Zeng et al. [77] proposed the *Counterfactual Generator*, which generates counterfactual examples for textual data and found that generated counterfactuals improved the generalizability of models under limited observational examples. Pitis et al. [56] proposed *Counterfactual Data Augmentation* (CoDA) for generating counterfactual experiences in reinforcement learning (RL), in which the method increases the size of available training data with counterfactual examples by stitching together locally-independent subsamples from the environment. They found that CoDA significantly improved the performance of RL agents in locally-factored tasks for batch-constrained and goal-conditioned settings. The problem with these papers is that they use bespoke counterfactual methods developed for specific task domains, rather than the tried and tested techniques from the XAI literature. Therefore, their performance, robustness and generalizability is, at best, uncertain[1].

However, one study has applied to a well-tested XAI counterfactual method to the problem of data augmentation. Temraz, et al. [67] used the present instance-guided counterfactual method to generate synthetic data for a crop-growth prediction problem. Their problem domain involved a *k*-NN model –called PBI-CBR -- for grass growth prediction that relies on an historical dataset of specific measurements of climate and grass growth on dairy farms in Ireland (N=70,091 data-points covering 2013-2018). The PBI-CBR model does reasonably well at predicting grass growth for individual farms in the coming week using this historical data. But, with climate change, there are an increasing number of climate-disruptive events, events that diverge significantly from the scenarios recorded in the historical data (e.g., extreme values for key weather variables like solar radiation or soil moisture). For example, in 2018 there was a significant drought across Europe, that effectively halted grass growth in Ireland during, what is usually, the peak-month of July (i.e., if soil moisture drops, then grass stops growing, indeed high solar radiation will burn grass). Accordingly, the PBI-CBR model does not do very well at predicting grass growth for these climate-disrupted months of 2018 because they are historically unique. Temraz et al. defined a *climate-based class boundary* in the PBI-CBR dataset, creating a division between "normal cases" (with climate-variable values within 2 standard deviations of historical means) and "climate-disrupted cases" (with climate-variable values >2 standard deviations of historical means). From a classification perspective, these "normal cases" were the majority class and the "climate-disrupted cases" are the minority class. They then used the instance-based counterfactual method to create new synthetic climate-disrupted cases and showed that the PBI-CBR model's performance specifically improved on predicting climate-disruptive events (in 2018) using these newly-created minority data-points. Interestingly, Temraz et al's experiments showed that that the instance-guided methods did better than optimization-methods in this problem domain (specifically, the DICE method [52]). However, this work only considers one specific problem domain, classifier, and dataset. It remains to be seen whether this counterfactual approach generalizes to other problem domains, classifiers and datasets; and, specifically, to datasets where class-imbalance problems arise. Hence, this is the aim for the remainder of this paper.

---

[1] An unpublished paper [48] reports an identical method to Wachter et al.'s. [70] counterfactual optimization method for data augmentation; however, this paper does not reference Wachter et. al. or any of the XAI literature.

## 3  The Counterfactual Augmentation Algorithm (CFA)

This paper advances the use of counterfactual methods for data augmentation as a solution to populating the minority class with more synthetic data to solve class-imbalance problems. The application of this XAI method to data augmentation was motivated by the observation that it seemed to generate plausible, synthetic datapoints for explanatory purposes; furthermore, the evaluation metrics in XAI showed that these datapoints were generally valid, within-distribution and close to existing datapoints. Accordingly, the extension of these techniques to data augmentation problems seemed like a promising avenue of research.

In this section, we describe a new oversampling method using a case-based reasoning approach to generating synthetic counterfactuals in the minority class, to be applied to binary classification problems. Consider a simple scenario to show how this method operates.

## 3.1 A Counterfactual Example for Class Imbalances

Imagine an ML classifier being used to predict whether farm animals are likely to be healthy or fall ill (e.g., mastitis in cows; see [60]). The dataset recording a herd of cows on most farms will be imbalanced, in that most cows will tend to be healthy rather than ill. An analysis of this dataset shows that some pairs of instances, majority-minority instance pairs, be counterfactually related to one another; for example, a majority instance, Cow-A of a certain breed, age, milk-yield and health-history that is classed as healthy can be counterfactually paired with, a minority instance, Cow-B that is of the same breed, age, milk-yield but with a different health-history (e.g., they have ill several times) that is classed as (likely to be) ill. This counterfactual-pair (which we call a *native counterfactual)* tells us that a one-feature difference in the health-history feature can change the class of a cow from healthy to ill. So, if we want to fix the class imbalance in this dataset using our counterfactual method, then one can generate a new minority instance by re-using this known counterfactual relationship. Imagine, we pick another majority instance, Cow-A' (a disease-free cow that has no existing counterfactual pair) and find a nearest neighbour to Cow-A' that is part of known counterfactual-pair (e.g., the Cow-A~Cow-B pair). Using this majority-instance and the native counterfactual-pair, we can generate a new, synthetic minority instance, using the match-features of Cow-A' and the difference-features from Cow-B; so, this new instance, Cow-B', would have the matching-features of Cow-A' (for breed, age and milk-yield) and the difference-feature from Cow-B (for health-history), along with the prediction that it will be ill. So, we have now created a new minority instance, Cow-B', that is counterfactually related to Cow-A' (n.b., where the class of this new instance needs to be verified by the underlying ML model)[2]. This example describes the generation of one minority instance. In our experiments, we do this iteratively for all those majority instances that are not paired in native counterfactuals; a step that results in the generation of many more minority instances close to the decision boundary with the majority class. In the next sub-section, we describe the algorithm more formally.

---

[2] Keane & Smyth [39] have shown good native counterfactual-pairs (i.e., those with >=2 differences between them) are quite rare in most datasets (<1% of all instances), but with some tolerance in feature-matching they can be increased (to ~5%).

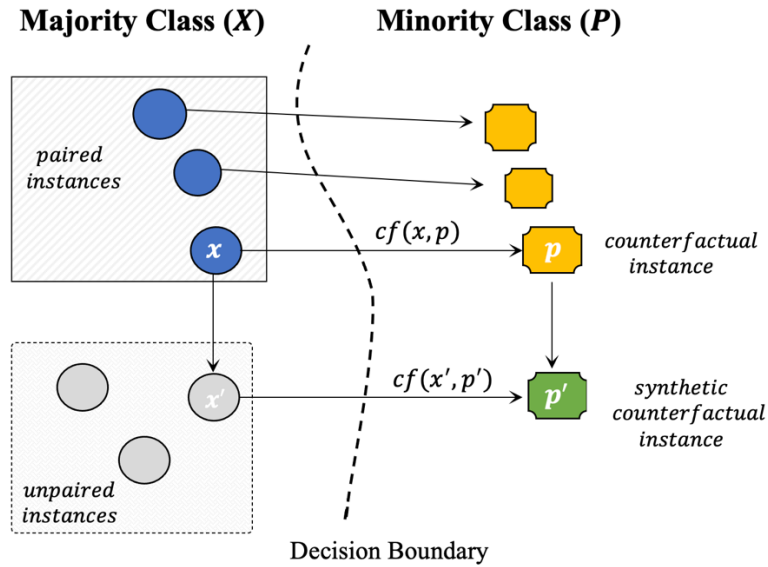

**Figure 1:** Counterfactual Augmentation (CFA): An unpaired instance, $x'$ (grey circle), finds a nearest neighbor, $x$ (blue circle), taking part in a "good" native counterfactual-pair in the dataset, $cf(x, p)$ (pairing of blue circle and yellow box) and then uses the difference-features of the counterfactual-instance, $p$ (yellow box), to generate a new synthetic counterfactual-instance, $p'$ (green box), combining them with the matching-features of the original unpaired instance, $x'$ (grey circle). The generated synthetic instance, $p'$ (green box), is then added to the dataset to improve future prediction.

### 3.2 The Method: Counterfactual Augmentation (CFA)

The *Counterfactual Augmentation* (CFA) is a technique for generating synthetic examples in the minority class using counterfactual methods (see [39, 64]) for the method used in XAI). The CFA method generates synthetic counterfactual instances in three main steps (see Figure 1):

i. "good" *native* counterfactuals, $cf(x,p)$, are initially computed over the whole dataset, $T$, identifying combinations of *counterfactually-paired* instances, $x$, from the majority class and, $p$, from the minority class

ii. given an *unpaired* instance, $x'$, its nearest-neighboring *paired* instance, $x$, is found and used to identify a close, existing counterfactual pair, $cf(x,p)$, and then

iii. a new, synthetic counterfactual instance, $p'$, is produced in the minority class using features from the original *unpaired* instance, $x'$ and feature-difference values from $p$.

More formally:

**Definitions**

- $X$ is a majority class, $(class_{maj})$
- $P$ is a minority class, $(class_{min})$
- $X = \{x, x'\}$, where:
  - $x$ is a paired instance in $class_{maj}$, $x = x_i\ (x_1, x_2, x_3, \dots, x_i)$, where $x \in cf(x,p)$
  - $x'$ is an unpaired instance in $class_{maj}$, $x' = x'_i\ (x'_1, x'_2, x'_3, \dots, x'_i)$, where $x' \notin cf(x,p)$
- $P = \{p, p'\}$, where:
  - $p$ is a counterfactual instance in $class_{min}$, $p = p_i\ (p_1, p_2, p_3, \dots, p_i)$, where $p \in cf(x,p)$
  - $p'$ is a synthetic counterfactual instance generated to be added to $class_{min}$
- CF-Set is $cf(x,p) \Longleftrightarrow target(x_i) \neq target(p_i)$
- $K$-nearest neighbors = $k$-NN

Assume that $x$ is a paired instance (which belongs to the *majority* class, $X$), and $p$ is a *counterfactual* instance (which belongs to the *minority* class, $P$):

$$x = \{x_1, x_2, x_3, \dots, x_i\}, \quad p = \{p_1, p_2, p_3, \dots, p_i\}$$

The procedure for CFA is as follows:

**Step 1** **Compute the CF-Set for the dataset, $cf(x, p)$:** CFA first finds all possible "good" *native* counterfactual pairs, $cf(x, p)$, between instances that already exist in a dataset, $T$; these native counterfactuals pair an instance in the *majority* space (called the *paired* instance) and its *counterfactually-related* instance in the *minority* space (called the *counterfactual* instance). In other words, for every $x_i$ in the majority class $x$, we find its counterfactual $p_i$ from $p$ in the minority class. These *native counterfactuals*, $cf(x, p)$, pair instances either side of *decision boundary* (they are called *native* because, in one sense, they already exist in the dataset). Each of these native pairs has a set of *match-features* and a set of *difference-features*, where the *differences* determine the class change over the decision boundary.

**Step 2** **For each unpaired instance, $x'$, from the majority class, find its nearest-neighbor paired instance, $x$, taking part in a native counterfactual, $cf(x, p)$:** For each unpaired instance, $x'$, CFA uses a *k*-NN to find its nearest neighboring, $x$, a paired instance involved in a native counterfactual pair, $cf(x, p)$. By definition, $x'$, belongs to the majority class and does not occur any native counterfactual pairs; notably, this means that all the synthetic datapoints generated by CFA come from these instances in the majority class that are *not* already counterfactually-related to instances in the minority class. Euclidean distance is used in finding these nearest neighbors:

$$\text{Euclidean Distance (ED)} = \sqrt{\sum_{i=1}^{m} (p_i - q_i)^2} \qquad (4)$$

**Step 3    Transfer feature-values from $p$ to $p'$ and from $x'$ to $p'$:** Having identified a candidate native counterfactual, $cf(x, p)$, for $x'$, CFA generates a synthetic counterfactual instance in the minority class, $p'$, using feature-values from $x'$ and $p$, such that:

- For each of the *difference-features* between $x$ and $p$, take the values from $p$ into the synthetic counterfactual case, $p'$.

- For each of the *match-features* between $x$ and $p$, take the values from $x'$ into the new counterfactual case, $p'$.

It should be noted that, tolerance is one parameter in CFA algorithm, which is used to improve the availability of good native counterfactuals in the dataset. Without tolerance, fewer counterfactuals would be found and the generative benefits of them would likely diminish. In finding matching- and difference-features between two instances for a native counterfactual, CFA computes a tolerance by finding the mean ($\mu$) and standard deviation ($\sigma$) for each feature. Then it allows features to match if their values are within +/-10% of the standard deviation from the mean all the values for that feature.

Two subtle differences that distinguish this data augmentation version of the algorithm from its XAI counterpart. First, although both algorithms adopt the same definition of a "good" counterfactual pairing the do so for different reasons. On psychological grounds, Keane and Smyth [39] defined a "good" counterfactual to be one with no more than two feature-differences. From the XAI perspective, researchers argue that "sparse" counterfactuals (with fewer feature differences) are better, because people find them more understandable (confirmed by user studies [22, 23]). From a data augmentation perspective, basing synthetic counterfactuals on sparse pairs also makes sense because the implicit causal dependencies between matched and difference features are more likely to be preserved in sparse-pairs; hence, generated synthetic data-points using these sparse pairs should be more likely to be valid and within-distribution. Second, there is a critical difference between the XAI and data augmentation contexts with respect to the selection of test instances. In XAI, the test instance is typically a novel problem for which a classifier has made a prediction, a prediction that needs to be counterfactually explained. Hence, typically, the test instance is not already in the training data. In data augmentation, the test instances used to generate synthetic counterfactual-data *have to be* in the training dataset; specifically, they are all the majority class instances in the training data that do not take part in native counterfactuals; this is why the "test instances" are called *unpaired instances*. For data augmentation purposes, the "test instances" are a residual set of majority instances (left after the native counterfactual-pairs have been identified).

## 3.3 How Counterfactuals Differ from SMOTE Variants

It should be apparent that this instance-based counterfactual method is quite different from SMOTE and its variants, though it is consistent with many insights from the class-imbalance literature. First, by definition, the counterfactual method addresses regions close to the decision boundary; a good counterfactual records the minimal feature-changes that result in a class change (as in Borderline-SMOTE and SVM-SMOTE). Second, this instance-based method relies on native counterfactuals in the dataset, pairings between existing majority and minority instances and, as such, is exploiting relationships between both classes (e.g., as in ADASYN, SMOTE-RSB, SL-SMOTE). Third, we are highly selective in the minority instances used to generate synthetic instances (as in the many clustering-driven SMOTE variants); that is, we only work of those involved in known counterfactuals with two feature differences). However, this counterfactual method is quite dif-ferent in many other significant respects. First, it does not use interpolation between majority/minority instances but rather uses the native-counterfactual as a template for generating new minority instances, (ii) it does not rely on the topology of the majority class (e.g., as in SWIM), but acts in a very local way using the counterfactual relation between a single majority instance and a minority one, (iii) does not rely on any clustering analysis of the majority and minority classes. As such, it represents quite a novel departure relative to existing SMOTE variants.

## 4  Competitive Tests of Data Augmentation Methods

In the current study, we competitively test the instance-based counterfactual method (CFA) against the benchmark techniques in the class-imbalance literature using six oversampling methods: SMOTE [11], B-SMOTE [27], ADASYN [30], SL-SMOTE [9], SVM-SMOTE [54], SMOTE-RSB [59]. These specific methods were chosen based of their conceptual closeness to the CFA method, their popularity amongst SMOTE variants and their public-availability as implementations. The six techniques were tested on a representative selection of 10 commonly-used UCI/KEEL datasets [2, 3, 4, 8], from which 25 dataset-variants were produced, with four different ML classifiers: including, Random Forest (RF), $k$-nearest neighbor ($k$-NN), Logistic Regression (LR), and Multilayer Perceptron (MLP) models. Several alternative ML models were used because different models find different decision boundaries for a given dataset, differences that could impact the success of the counterfactual method (as it relies heavily on a model's decision boundary). The Baseline-Control for a given classifier recorded the performance of the model on a given dataset without any data augmentation applied. Several standard measures were used to assess the performance of the four methods; namely, Precision, Recall, F1, and plots of ROC curves.

## 4.1 Method: Datasets & Setup

Table 2 shows the main characteristics of the datasets drawn from both UCI and KEEL repositories. As the focus is on binary classification problems and some of these datasets are multi-class, they were converted to binary classes. The One-Versus-One (OVO or 1v1) [7, 33] and One-Versus-Rest (OVR or 1vR) [7, 13, 69] methods were used to do this conversion. The OVO method splits a multi-class classification into one binary classification dataset for each pair of classes. Whereas the OVR approach selects one of the multiple classes and predicts it against all other classes. So that one of the classes is treated as the positive (minority) class, and all other classes are treated as the negative class (majority). In this paper, the datasets were modified using both methods (one method per dataset) to vary the class ratio of class imbalance among the datasets (see Table 2). The 10 base datasets were:

- *Abalone dataset:* A multi-class dataset analyzed to find the age of abalone from physical measurements, consisting of 28 classes, that was modified using 1v1 and 1vR.

- *Glass dataset:* A multi-class dataset used to classify glass-types based on the chemical analysis, consisting of 7 classes, that modified using 1vR to treat the class '3' as the minority class, and all other classes are treated as the negative class.

- *Yeast dataset:* A multi-class dataset used to predict the cellular localization sites of proteins, consisting of 10 classes, that modified using 1v1 and 1vR.

- *Pima Indians Diabetes dataset:* A binary-class dataset used to predict whether or not a patient has diabetes, based on certain diagnostic measurements included in the dataset.

- *Phoneme dataset:* A binary-class dataset used to distinguish between nasal and oral sounds.

- *Vehicle dataset:* The Vehicle data set is a multi-class dataset, with 4 classes. The problem is to classify a given silhouette as one of four types of vehicle, using a set of features extracted from the silhouette. Again, this dataset was modified using 1vR. So that the class "Van" is treated as the minority class, and all other classes are treated as the negative class.

- *Ecoli dataset:* This dataset is a multi-class dataset, with 8 classes. The problem is to classify Ecoli proteins using their amino acid sequences in their cell localization sites.

- *Page-Blocks dataset:* A multi-class dataset used to classify all the blocks of the page layout of a document that has been detected by a segmentation process.

- *Wine Quality dataset (Red and White):* Two datasets related to red and white vinho verde wine samples. The problem is to classify wine quality based on physicochemical tests.

- *Poker dataset:* A multi-class dataset, with 10 classes used to predict poker hands.

The overall performance of each classifier was tested using $k$-fold cross validation (with $k = 5$). So, each dataset is randomly partitioned into 5 disjoint subsets, where each subset included approximately equal size of data; then, a single subset was retained as a test set with the remaining $k$-1 subsets being used as the training data. Different datasets have different-sized minority classes (see Table 2 for details). For each of the SMOTE methods, we split each dataset into training and validation folds. Then, on each fold we oversample the minority class, train the classifier on the training folds, and finally, validate the classifier on the remaining fold. For CFA, the native counterfactuals in the training data were computed (CF-Set) and then all the remaining unpaired majority-instances were run through CFA to create the synthetic counterfactual instances in the minority class; this augmented dataset was then used for testing. These generated datasets from CFA were compared with the original dataset (without data augmentation) and the datasets generated by SMOTE, B-SMOTE, ADASYN, SVM-SMOTE, SL-SMOTE and SMOTE-RSB. For our experiments, we oversample the minority class using each of the data augmentation methods until we have the same number of instances in each class. As a result, fully balanced datasets were created. For each classifier based on the original imbalanced dataset was also run as a baseline, without using any data augmentation method. Finally, to determine the optimal values for all classifiers, we applied hyperparameters tuning using GridSearchCV function from *scikit-learn*. GridSearchCV performs cross-validated grid-search across all hyperparameter combinations and finds the best score for a given classifier. To achieve this, we defined our grid of parameters for each classification method (RF, *k*-NN, LR, MLP) and oversampling methods (SMOTE & its variants) and then ran the grid search (see Table 1 for a full description).

**Table 1:** A grid of hyperparameter values for each classifier and oversampling methods

| Learner | Parameter Variants |
|---|---|
| Random Forest (RF) | $ntree = \{50, 100, 200, 400, 600\}$, $max_{depth} = \{4, 6, 10, 20, 30, 50, 80, 100\}$ |
| *k*-nearest neighbors (*k*-NN) | $n_{neighbors} = \{3, 5, 7, 10, 20, 30, 50\}$ |
| Logistic Regression (LR) | $max_{iter} = \{1000, 200\}$, $C = \{0.001, 0.01, 0.1, 1, 10, 100, 1000\}$, $solver = \{'newton-cg', 'lbfgs', 'liblinear', 'sag'\}$ |
| Multilayer Perceptron (MLP) | $hidden_{layer\,sizes} = \{(10, 10, 10), (10,30,10), (50,50), (100,50)\}$, $activation = \{'tanh', 'relu'\}$, $alpha = \{0.0001, 0.05\}$, $solver = \{'sgd', 'adam'\}$ |
| Oversampling methods (SMOTE & its variants) | $k_{nearest\,neighbors} = \{3, 5, 7, 9, 20\}$ |

Table 2: Datasets & DataSet Variants Used in the Experiment (IR= Imbalance Ratio)

| ID | Dataset | Features | Instances | Minority | Majority | IR |
|---|---|---|---|---|---|---|
| D1 | Pima | 9 | 768 | 268 | 500 | 1.86000 |
| D2 | Phoneme | 6 | 5404 | 1586 | 3818 | 2.40000 |
| D3 | Vehicle | 19 | 846 | 199 | 647 | 3.25000 |
| D4 | Abalone-9-vs-13 | 9 | 892 | 203 | 689 | 3.39000 |
| D5 | Yeast-3-vs-R | 9 | 1484 | 163 | 1321 | 8.10000 |
| D6 | Ecoli-3-vs-R | 8 | 336 | 35 | 301 | 8.60000 |
| D7 | Page-Blocks-0-vs-R | 11 | 5472 | 559 | 4913 | 8.78000 |
| D8 | Yeast-0-3-5-9-vs-7-8 | 9 | 506 | 50 | 456 | 9.12000 |
| D9 | Abalone-9-vs-16 | 9 | 756 | 67 | 689 | 10.2800 |
| D10 | Glass-3-vs-R | 10 | 214 | 17 | 197 | 11.5800 |
| D11 | WineQuality-Red-4-vs-5 | 12 | 734 | 53 | 681 | 12.8400 |
| D12 | Yeast-1-vs-7 | 9 | 459 | 30 | 429 | 14.3000 |
| D13 | Ecoli-4-vs-R | 8 | 336 | 20 | 316 | 15.8000 |
| D14 | Abalone-13-vs-R | 9 | 4177 | 203 | 3974 | 19.5700 |
| D15 | Abalone-9-vs-19 | 9 | 721 | 32 | 689 | 21.5300 |
| D16 | Abalone-9-vs-20 | 9 | 715 | 26 | 689 | 26.5000 |
| D17 | Yeast-4-vs-R | 9 | 1484 | 51 | 1433 | 28.0900 |
| D18 | WineQuality-Red-6-vs-8 | 12 | 656 | 18 | 638 | 35.4400 |
| D19 | Abalone-17-vs-7-8-9-10 | 9 | 2338 | 58 | 2280 | 39.3100 |
| D20 | Yeast-6-vs-R | 9 | 1484 | 35 | 1449 | 41.4000 |
| D21 | WineQuality-White-3-vs-7 | 12 | 900 | 20 | 880 | 44.0000 |
| D22 | WineQuality-White-3-9-vs-5 | 12 | 1482 | 25 | 1457 | 58.2800 |
| D23 | Poker-8-9-vs-6 | 11 | 1485 | 25 | 1460 | 58.4000 |
| D24 | Abalone-20-vs-8-9-10 | 9 | 1916 | 26 | 1890 | 72.6900 |
| D25 | Poker-8-9-vs-5 | 11 | 2075 | 25 | 2050 | 82.0000 |

## 4.2 Metrics & Measures

In binary classification problems, the labels can be either positive or negative. So, the prediction made by the classifier is represented as a 2 × 2 confusion matrix [35] (see Table 3). The confusion matrix summarizes the performance of classifiers for the four possible outcomes of a given classification: a true positive (TP), true negative (TN), false positive (FP) and false negative (FN). Accuracy was not used as a measure because, as discussed earlier, it can be spuriously high for imbalanced datasets. It should be noted that all datasets used in our experiments were converted to binary datasets using two of the most well-known strategies; 1v1 [7, 33] and 1vR [7, 13, 69] (see section 4.1 for a full description). Hence, the evaluation metrics used were as *precision*, *recall*, AUC and *F1* defined as follows:

$$Recall = \frac{TP}{(TP + FN)} \quad (5)$$

$$Precision = \frac{TP}{(TP + FP)} \quad (6)$$

$$F_1 = 2 * \frac{(precision * recall)}{(precision + recall)} \quad (7)$$

$$AUC = \frac{(TP + TN)}{2} \quad (8)$$

**Table 3:** Confusion Matrix for Classifications

|  |  | Predicted Class | |
|---|---|---|---|
|  |  | P | N |
| Actual Class | P | True Positive (TP) | False Negative (FN) |
|  | N | False Positive (FP) | True Negative (TN) |

Receiver Operating Characteristic curves (ROC curve) were reported as they are often used to evaluate classification models for imbalanced data sets [26]. The ROC curve is a two-dimensional graph in which true positive (TP) rate is plotted on the y-axis and false positive (FP) rate is plotted on the x-axis. One advantage of ROC curves is that they are not affected by the class ratio between minority and majority instances in the datasets (see Figures for results). Area Under Curve (AUC) scores were also reported as they are used to measure the two-dimensional area that lies under the ROC curve.

### 4.3. Results & Discussion

Overall, the counterfactual data-augmentation method (CFA) performs better than all other SMOTE-based methods on the main metrics reported, for most of the classifiers tested (see Tables 4-7). Recall, the cross-validated datasets were tested on four classifiers (RF, *k*-NN, LR, MLP) with the Baseline (no data augmentation), SMOTE, B-SMOTE, ADASYN, SVM-SMOTE, SL-SMOTE, SMOTE-RSB and CFA. Tables 4-7 report the main metric (AUC-ROC) for each classifier on the datasets. For the *RF classifier* (Table 4), the results show that CFA does better than all the other SMOTE-based methods in 21 datasets out of 25, with SVM-SMOTE being the next best in only 2 datasets. Both ADASYN and SL-SMOTE had the highest AUC-ROC for one dataset for each. For the *k-NN classifier* (Table 5), it is observed that, CFA also achieved a greater improvement in AUC-ROC. The results also show that for AUC-ROC, CFA does better than all the other SMOTE-based methods in 19 out of 25 datasets, with ADASYN being the next best with 2 datasets. SMOTE, B-SMOTE, SVM-SMOTE and SMOTE-RSB had the highest AUC-ROC in 1 dataset for each method. For the *LR classifier* (Table 6), the results are quite different in that they show that for AUC-ROC metric, CFA doing better in only 10 out of 25 datasets, with the SMOTE-RSB being the next best with 5 datasets. Whereas Baseline, B-SMOTE, ADASYN, SVM-SMOTE and SL-SMOTE doing better in 2 datasets. Finally, for the *MLP classifier* (Table 7), again the results showed that CFA does better than all the other SMOTE-based methods in 16 out of 25 datasets. Whereas, SMOTE-RSB doing better in only 5 datasets, with SVM-SMOTE being the next best with 3 datasets. Baseline, B-SMOTE, and ADASYN had the highest AUC-ROC in 2 datasets for each method. Notably, these results show for certain datasets and classifiers, the SMOTE-RSB does quite well; However, when data augmentation can make a contribution, it seems to be CFA that contributes the most to performance improvements.

Table 4: AUC values for the RF classifier for each Data Augmentation Method

| Dataset | Baseline | SMOTE | B-SMOTE | ADASYN | SVM-SMOTE | SL-SMOTE | SMOTE-RSB | CFA |
|---|---|---|---|---|---|---|---|---|
| D1 | 0.8519 | 0.8505 | 0.8461 | 0.8467 | 0.8496 | 0.8483 | 0.8540 | **0.9070** |
| D2 | 0.9580 | 0.9593 | 0.9577 | 0.9594 | 0.9583 | 0.9507 | 0.9585 | **0.9766** |
| D3 | 0.9931 | 0.9938 | 0.9940 | 0.9935 | 0.9941 | **0.9942** | 0.9941 | 0.9847 |
| D4 | 0.8124 | 0.8133 | 0.8214 | 0.8213 | 0.8222 | 0.7959 | 0.8120 | **0.9121** |
| D5 | 0.9693 | 0.9732 | 0.9708 | 0.9741 | 0.9719 | 0.9424 | 0.9764 | **0.9948** |
| D6 | 0.9519 | 0.9640 | 0.9672 | 0.9664 | 0.9656 | 0.9519 | 0.9594 | **0.9896** |
| D7 | 0.9906 | 0.9900 | 0.9898 | 0.9891 | 0.9907 | 0.9817 | 0.9906 | **0.9974** |
| D8 | 0.7869 | 0.7943 | 0.7916 | 0.8002 | 0.7932 | 0.7728 | 0.8177 | **0.9373** |
| D9 | 0.8897 | 0.9136 | 0.9153 | 0.9169 | 0.9074 | 0.8572 | 0.8915 | **0.9939** |
| D10 | 0.8541 | 0.8774 | 0.9062 | 0.8797 | 0.8945 | 0.8568 | 0.8582 | **0.9802** |
| D11 | 0.7874 | 0.7477 | 0.7717 | 0.7370 | 0.7630 | 0.6249 | 0.7954 | **0.9789** |
| D12 | 0.8489 | 0.8414 | 0.8717 | 0.8529 | 0.8596 | 0.7647 | 0.8877 | **0.9827** |
| D13 | 0.9915 | 0.9944 | 0.9944 | 0.9951 | 0.9986 | 0.9915 | 0.9986 | **0.9992** |
| D14 | 0.7497 | 0.7549 | 0.7593 | 0.7627 | 0.7626 | 0.6682 | 0.7499 | **0.9801** |
| D15 | 0.8861 | 0.9119 | 0.9129 | 0.9146 | 0.9086 | 0.7954 | 0.9148 | **0.9904** |
| D16 | 0.9727 | 0.9728 | 0.9683 | **0.9734** | 0.9652 | 0.9347 | 0.9685 | 0.9727 |
| D17 | 0.9145 | 0.9233 | 0.9310 | 0.9255 | 0.9350 | 0.8568 | 0.9260 | **0.9959** |
| D18 | 0.8789 | 0.9234 | 0.9386 | 0.9197 | 0.9203 | 0.8718 | 0.8939 | **0.9978** |
| D19 | 0.9241 | 0.9411 | 0.9455 | 0.9440 | **0.9484** | 0.9241 | 0.9232 | 0.9241 |
| D20 | 0.9546 | 0.9589 | 0.9525 | 0.9546 | 0.9651 | 0.9022 | 0.9592 | **0.9962** |
| D21 | 0.9227 | 0.8598 | 0.9247 | 0.8737 | 0.9040 | 0.9306 | 0.9505 | **0.9986** |
| D22 | 0.8058 | 0.8047 | 0.8166 | 0.7988 | 0.8293 | 0.8218 | 0.8394 | **0.9993** |
| D23 | 0.9588 | 0.9566 | 0.9710 | 0.9846 | **0.9901** | 0.9716 | 0.9766 | 0.9801 |
| D24 | 0.9375 | 0.9666 | 0.9728 | 0.9684 | 0.9697 | 0.9184 | 0.9388 | **0.9987** |
| D25 | 0.7708 | 0.7978 | 0.7265 | 0.8045 | 0.8286 | 0.7531 | 0.8118 | **0.9888** |
| Total | 0 | 0 | 0 | 1 | 2 | 1 | 0 | **21** |

Table 5: AUC values for the *k*-NN classifier for each Data Augmentation Method

| Dataset | Baseline | SMOTE | B-SMOTE | ADASYN | SVM-SMOTE | SL-SMOTE | SMOTE-RSB | CFA |
|---|---|---|---|---|---|---|---|---|
| D1 | 0.8247 | 0.8230 | 0.8232 | **0.8257** | 0.8238 | 0.8201 | 0.8254 | 0.7948 |
| D2 | 0.9255 | 0.9326 | 0.9285 | 0.9341 | 0.9309 | 0.9016 | 0.9335 | **0.9408** |
| D3 | 0.9773 | 0.9797 | 0.9814 | 0.9801 | **0.9818** | 0.9432 | 0.9776 | 0.9725 |
| D4 | 0.7858 | 0.7934 | 0.7884 | 0.7999 | 0.7818 | 0.7336 | 0.7871 | **0.8773** |
| D5 | 0.9699 | 0.9734 | 0.9645 | 0.9700 | 0.9678 | 0.8953 | 0.9693 | **0.9943** |
| D6 | 0.9639 | 0.9645 | 0.9597 | 0.9632 | 0.9646 | 0.9639 | **0.9669** | 0.9648 |
| D7 | 0.9574 | 0.9706 | 0.9674 | 0.9717 | 0.9683 | 0.8672 | 0.9578 | **0.9844** |
| D8 | 0.7737 | 0.7869 | 0.7721 | 0.7862 | 0.7946 | 0.7393 | 0.7767 | **0.8620** |
| D9 | 0.8220 | 0.8968 | 0.8900 | 0.8973 | 0.9124 | 0.7043 | 0.8663 | **0.9822** |
| D10 | 0.7648 | 0.7562 | 0.7792 | 0.7551 | 0.7666 | 0.7541 | 0.7180 | **0.9636** |
| D11 | 0.7122 | 0.6960 | 0.7105 | 0.6901 | 0.7216 | 0.5144 | 0.7055 | **0.9070** |
| D12 | 0.8406 | 0.8590 | 0.8460 | 0.8443 | 0.8501 | 0.5970 | 0.8428 | **0.9368** |
| D13 | 0.9979 | 0.9986 | 0.9986 | 0.9986 | 0.9986 | 0.9979 | 0.9986 | **0.9988** |
| D14 | 0.7468 | 0.7432 | 0.7336 | 0.7394 | 0.7304 | 0.5741 | 0.7359 | **0.9602** |
| D15 | 0.8248 | 0.8806 | 0.8821 | 0.8776 | 0.8714 | 0.6305 | 0.8878 | **0.9651** |
| D16 | 0.8271 | **0.8888** | 0.8657 | 0.8816 | 0.8463 | 0.6940 | 0.8244 | 0.8271 |
| D17 | 0.9054 | 0.9240 | 0.9235 | 0.9170 | 0.9230 | 0.7035 | 0.9115 | **0.9801** |
| D18 | 0.8285 | 0.8609 | 0.8494 | 0.8549 | 0.8440 | 0.7290 | 0.8468 | **0.9774** |
| D19 | 0.8409 | 0.8947 | 0.8960 | **0.8992** | 0.8908 | 0.8409 | 0.8465 | 0.8409 |
| D20 | 0.9241 | 0.9482 | 0.9235 | 0.9420 | 0.9244 | 0.8579 | 0.9209 | **0.9642** |
| D21 | 0.6553 | 0.7896 | 0.7452 | 0.8230 | 0.7470 | 0.5957 | 0.8551 | **0.9859** |
| D22 | 0.6454 | 0.7051 | 0.7178 | 0.7037 | 0.7181 | 0.6440 | 0.7216 | **0.9805** |
| D23 | 0.9889 | 0.9999 | **1.0000** | 0.9999 | 0.9999 | 0.9274 | 0.9998 | 0.9110 |
| D24 | 0.8541 | 0.8936 | 0.8774 | 0.8936 | 0.8794 | 0.7591 | 0.8526 | **0.9916** |
| D25 | 0.6785 | 0.7878 | 0.6869 | 0.7722 | 0.7362 | 0.6680 | 0.7566 | **0.9276** |
| Total | 0 | 1 | 1 | 2 | 1 | 0 | 1 | **19** |

Table 6: AUC values for the LR classifier for each Data Augmentation Method

| Dataset | Baseline | SMOTE | B-SMOTE | ADASYN | SVM-SMOTE | SL-SMOTE | SMOTE-RSB | CFA |
|---|---|---|---|---|---|---|---|---|
| D1 | 0.8517 | 0.8499 | 0.8525 | 0.8512 | 0.8523 | 0.8530 | **0.8533** | 0.7879 |
| D2 | 0.8202 | 0.8190 | 0.8137 | 0.8150 | 0.8149 | **0.8215** | 0.8188 | 0.7306 |
| D3 | 0.9967 | **0.9970** | 0.9967 | 0.9969 | 0.9962 | 0.9912 | 0.9967 | 0.9764 |
| D4 | 0.8373 | 0.8411 | 0.8439 | 0.8429 | 0.8380 | 0.8434 | 0.8441 | **0.8891** |
| D5 | 0.9659 | 0.9661 | 0.9639 | 0.9664 | 0.9657 | 0.9392 | 0.9664 | **0.9915** |
| D6 | 0.9546 | 0.9572 | 0.9557 | 0.9546 | **0.9582** | 0.9546 | 0.9484 | 0.9223 |
| D7 | 0.9379 | 0.9559 | **0.9601** | 0.9596 | 0.9575 | 0.9457 | 0.9385 | 0.9507 |
| D8 | 0.7919 | 0.7965 | 0.7790 | 0.7962 | 0.7903 | 0.7300 | **0.8012** | 0.7880 |
| D9 | 0.9453 | 0.9479 | 0.9487 | 0.9511 | 0.9477 | 0.9239 | 0.9412 | **0.9897** |
| D10 | 0.6710 | 0.8866 | 0.8989 | 0.8889 | 0.8897 | 0.8439 | 0.8147 | **0.9400** |
| D11 | 0.7254 | 0.7289 | **0.7405** | 0.7274 | 0.7381 | 0.6331 | 0.6654 | 0.7308 |
| D12 | 0.8619 | 0.8647 | 0.8589 | 0.8634 | 0.8580 | 0.5303 | 0.8642 | **0.9215** |
| D13 | 0.9944 | 0.9944 | 0.9944 | 0.9972 | **0.9986** | 0.9944 | 0.9958 | 0.9973 |
| D14 | 0.7374 | 0.7459 | 0.7480 | 0.7474 | 0.7452 | 0.6963 | 0.7480 | **0.8384** |
| D15 | 0.9494 | 0.9576 | 0.9553 | **0.9583** | 0.9566 | 0.8843 | **0.9583** | 0.9255 |
| D16 | 0.9848 | 0.9836 | 0.9835 | 0.9837 | 0.9830 | 0.9391 | **0.9852** | 0.9848 |
| D17 | 0.8814 | 0.8853 | 0.8833 | 0.8854 | 0.8848 | 0.6824 | 0.8862 | **0.9222** |
| D18 | 0.9186 | 0.9158 | 0.9218 | 0.9179 | 0.9166 | 0.6831 | 0.9078 | **0.9399** |
| D19 | 0.9408 | 0.9440 | 0.9422 | **0.9454** | 0.9412 | 0.9408 | 0.9432 | 0.9408 |
| D20 | **0.9617** | 0.9611 | 0.9537 | 0.9600 | 0.9555 | 0.8709 | 0.9592 | 0.8918 |
| D21 | 0.7328 | 0.7581 | 0.7657 | 0.7571 | 0.7460 | 0.6823 | 0.7485 | **0.8522** |
| D22 | **0.7425** | 0.6720 | 0.6959 | 0.6771 | 0.6978 | 0.6714 | 0.7138 | 0.6923 |
| D23 | 0.5114 | 0.4395 | 0.4950 | 0.6028 | 0.5101 | 0.6143 | 0.4834 | **0.6417** |
| D24 | 0.9763 | 0.9682 | 0.9720 | 0.9683 | 0.9731 | 0.8686 | **0.9765** | 0.9657 |
| D25 | 0.5366 | 0.5365 | 0.4795 | 0.5456 | 0.4874 | **0.5574** | 0.5414 | 0.5504 |
| Total | 2 | 1 | 2 | 2 | 2 | 2 | 5 | 10 |

Table 7: AUC values for the MLP classifier for each Data Augmentation Method

| Dataset | Baseline | SMOTE | B-SMOTE | ADASYN | SVM-SMOTE | SL-SMOTE | SMOTE-RSB | CFA |
|---|---|---|---|---|---|---|---|---|
| D1 | 0.8535 | 0.8532 | 0.8512 | 0.8513 | **0.8550** | 0.8529 | 0.8545 | 0.8083 |
| D2 | 0.9194 | 0.9235 | 0.9185 | 0.9223 | 0.9211 | 0.9103 | **0.9251** | 0.8883 |
| D3 | 0.9984 | 0.9984 | 0.9986 | 0.9985 | **0.9987** | 0.9955 | 0.9985 | 0.9882 |
| D4 | 0.8279 | 0.8331 | 0.8382 | 0.8355 | 0.8354 | 0.8294 | 0.8452 | **0.9017** |
| D5 | 0.9741 | 0.9742 | 0.9723 | 0.9732 | 0.9736 | 0.9443 | 0.9761 | **0.9945** |
| D6 | 0.8955 | 0.9679 | **0.9691** | 0.9639 | 0.9666 | 0.8955 | 0.9432 | 0.9590 |
| D7 | 0.9839 | 0.9869 | 0.9849 | 0.9854 | 0.9868 | 0.9600 | 0.9834 | **0.9871** |
| D8 | 0.8019 | 0.7988 | 0.7831 | 0.7960 | 0.7936 | 0.7408 | 0.8078 | **0.8504** |
| D9 | 0.8837 | 0.9357 | 0.9375 | 0.9362 | 0.9346 | 0.9211 | 0.9347 | **0.9888** |
| D10 | 0.6583 | 0.7976 | 0.7997 | 0.8285 | 0.8713 | 0.7251 | 0.7277 | **0.9322** |
| D11 | 0.7107 | 0.7431 | 0.7476 | 0.7372 | 0.7484 | 0.6977 | 0.6997 | **0.8499** |
| D12 | 0.8589 | 0.8619 | 0.8508 | 0.8611 | 0.8548 | 0.7698 | 0.8594 | **0.9205** |
| D13 | 0.9776 | 0.9958 | 0.9972 | 0.9972 | 0.9972 | 0.9776 | 0.9958 | **0.9973** |
| D14 | 0.7554 | 0.7811 | 0.7752 | 0.7863 | 0.7737 | 0.6790 | 0.7804 | **0.9024** |
| D15 | 0.9478 | 0.9524 | 0.9491 | 0.9504 | 0.9503 | 0.8579 | **0.9535** | 0.9418 |
| D16 | 0.9831 | 0.9834 | 0.9849 | 0.9830 | 0.9858 | 0.9129 | **0.9873** | 0.9831 |
| D17 | 0.8924 | 0.8946 | 0.8901 | 0.8954 | 0.8849 | 0.8281 | 0.8969 | **0.9602** |
| D18 | 0.9033 | 0.9131 | 0.9158 | 0.9117 | 0.9141 | 0.7374 | 0.9040 | **0.9847** |
| D19 | 0.9301 | 0.9466 | 0.9454 | **0.9468** | 0.9463 | 0.9301 | 0.9464 | 0.9301 |
| D20 | 0.9635 | 0.9618 | 0.9595 | 0.9604 | 0.9594 | 0.8375 | **0.9652** | 0.9526 |
| D21 | 0.7768 | 0.8328 | 0.8030 | 0.8066 | 0.7702 | 0.7045 | 0.7525 | **0.9935** |
| D22 | 0.6063 | 0.6370 | 0.6550 | 0.6220 | 0.6884 | 0.5991 | 0.6999 | **0.9797** |
| D23 | **1.0000** | **1.0000** | **1.0000** | **1.0000** | **1.0000** | 0.9790 | **1.0000** | 0.9218 |
| D24 | 0.9659 | 0.9672 | 0.9690 | 0.9671 | 0.9682 | 0.8746 | 0.9643 | **0.9847** |
| D25 | 0.7453 | 0.7556 | 0.7663 | 0.7624 | 0.7262 | 0.6959 | 0.7425 | **0.9304** |
| Total | 2 | 1 | 2 | 2 | 3 | 0 | 5 | **16** |

Notably, if we assess overall performance by noting occasions for a given method when the Precision score is highest, we see that CFA scores best (see Table 8): in 70.0% (70/100) of cases it has the highest Precision score, as opposed to 24% (24/100) for SMOTE-RSB, 14% (14/100) for Baseline, 4% (4/100) for SL-SMOTE, 3% (3/100) for ADASYN, 2% (2/100) for SMOTE, B-SMOTE and SVM-SMOTE. It is also observed that, CFA scores best in 61.0% (61/100) of cases it has the highest Recall score, as opposed to 31.0% (31/100) for ADASYN, 22.0% (22/100) for both SMOTE and B-SMOTE, 18.0% (18/100) for SMOTE-RSB, 11.0% (11/100) for SVM-SMOTE, 4.0% (4/100) for SL-SMOTE and 0% (0/24) for Baseline.

**Table 8:** The number of datasets for each method showing the highest Precision and Recall scores for a given method (SMOTE, B-SMOTE, ADASYN, SVM-SMOTE, SL-SMOTE, SMOTE-RSB and CFA) on a selected classifier

| | Precision | | | | | | | |
|---|---|---|---|---|---|---|---|---|
| Classifier | Baseline | SMOTE | B-SMOTE | ADASYN | SVM-SMOTE | SL-SMOTE | SMOTE-RSB | CFA |
| RF | 2 | 1 | 1 | 1 | 1 | 1 | 8 | 16 |
| $k$-NN | 4 | 0 | 0 | 1 | 0 | 1 | 6 | 16 |
| LR | 4 | 0 | 0 | 0 | 0 | 1 | 5 | 17 |
| MLP | 4 | 1 | 1 | 1 | 1 | 1 | 5 | 21 |
| Total | **14** | **2** | **2** | **3** | **2** | **4** | **24** | **70** |
| | Recall | | | | | | | |
| Classifier | Baseline | SMOTE | B-SMOTE | ADASYN | SVM-SMOTE | SL-SMOTE | SMOTE-RSB | CFA |
| RF | 0 | 1 | 1 | 2 | 0 | 1 | 1 | 20 |
| $k$-NN | 0 | 4 | 6 | 5 | 3 | 0 | 4 | 15 |
| LR | 0 | 10 | 10 | 13 | 5 | 2 | 7 | 13 |
| MLP | 0 | 7 | 5 | 11 | 3 | 1 | 6 | 13 |
| Total | **0** | **22** | **22** | **31** | **11** | **4** | **18** | **61** |

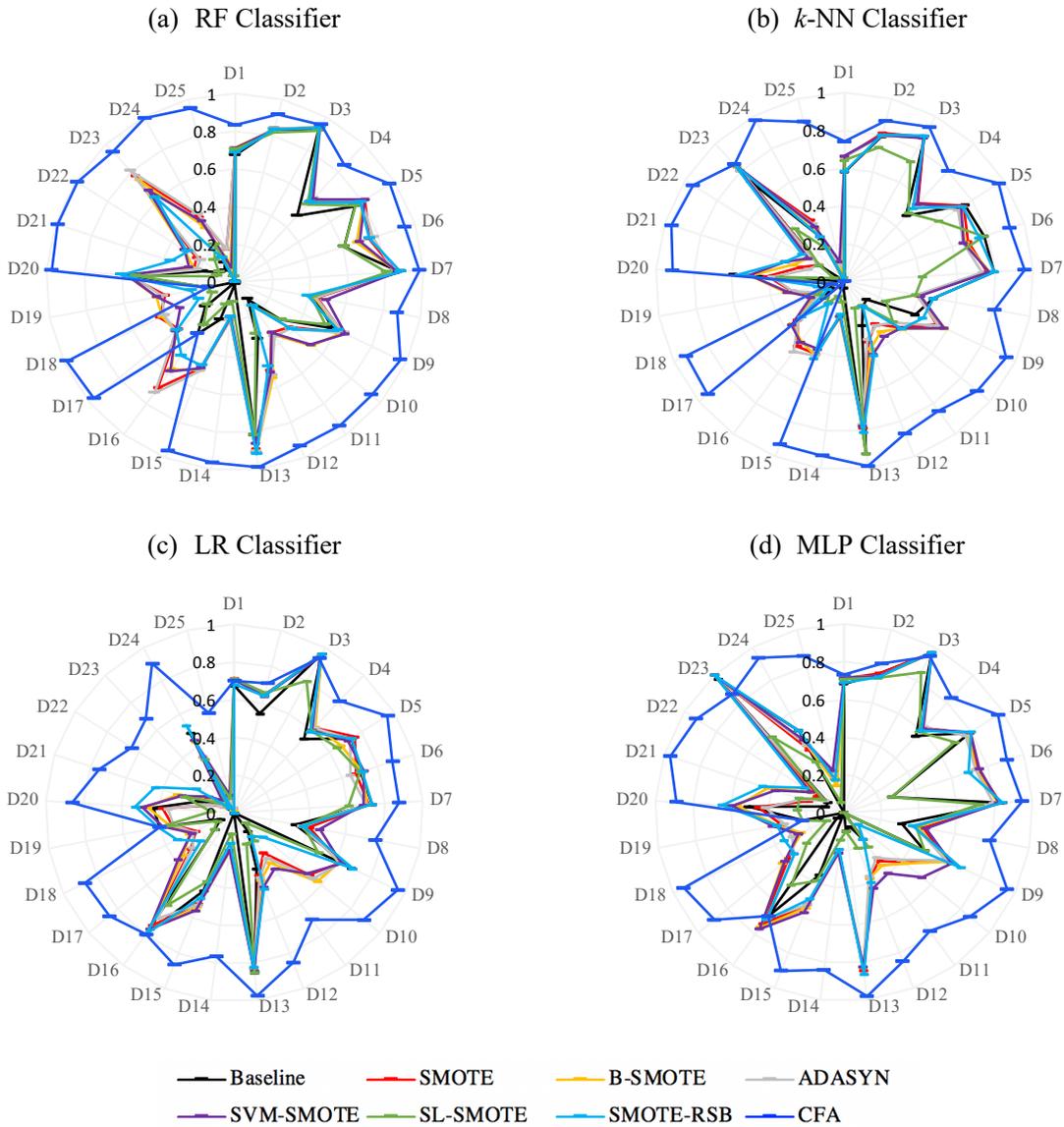

**Figure 2:** F1 values for the different conditions, across 25 datasets for the four classifiers (a) RF classifier, (b) *k*-NN, (c) LR, and (d) MLP

Figure 2 shows the F1 comparisons for each of the data augmentation methods (SMOTE, B-SMOTE, ADASYN, SVM-SMOTE, SL-SMOTE, SMOTE-RSB, and CFA) on 25 datasets using the four classifiers (RF, *k*-NN, LR, MLP). In general, higher F1 values indicate better classifier performance. Overall, there is no significant difference in the F1 values between SMOTE variants (SMOTE, B-SMOTE, ADASYN, SVM-SMOTE, SL-SMOTE, and SMOTE-RSB) and these algorithms achieved lower values than CFA in most cases. Perhaps the most interesting result is that CFA achieved higher F1 values in 23 out of 25 datasets when using the RF classifier and in 22 out of 25 datasets when using *k*-NN classifier. It outperformed the Baseline (with no data augmentation) and SMOTE variants. Similarly, when applying LR classifier, CFA achieved the best performance in 22 out of 25 datasets, although the improvement varied with different data sets. Finally, in the MLP classifier, CFA still achieved highest AUC in 21 out of 15 datasets.

Finally, the ROC curves, which show the trade-off between sensitivity and specificity, are presented in Figure 3 and 4. Figure 3 shows selected examples of ROC curves where CFA outperformed SMOTE-based methods, obtained for the four methods using the four classifiers on different data sets. According to Figure 3, CFA clearly outperformed other data augmentation methods (i.e., SMOTE, B-SMOTE, ADASYN, SVM-SMOTE, SL-SMOTE, SMOTE-RSB). For example, when running a RF classifier on the 'Abalone-9-vs-13' dataset we can see that CFA had better performance than SMOTE-based methods with respect to ROC curve (Figure 3). These results support our previous results which were showing that CFA can work as a successful technique for data augmentation to handling the class imbalanced problem. On the other hand, Figure 4 shows several examples of ROC curves where SMOTE variants (i.e., SMOTE, B-SMOTE, ADASYN, SVM-SMOTE, SL-SMOTE, SMOTE-RSB) do better than CFA. For example, when running a LR classifier on the 'PIMA' dataset we can see that SMOTE-based methods had better performance than CFA (Figure 4).

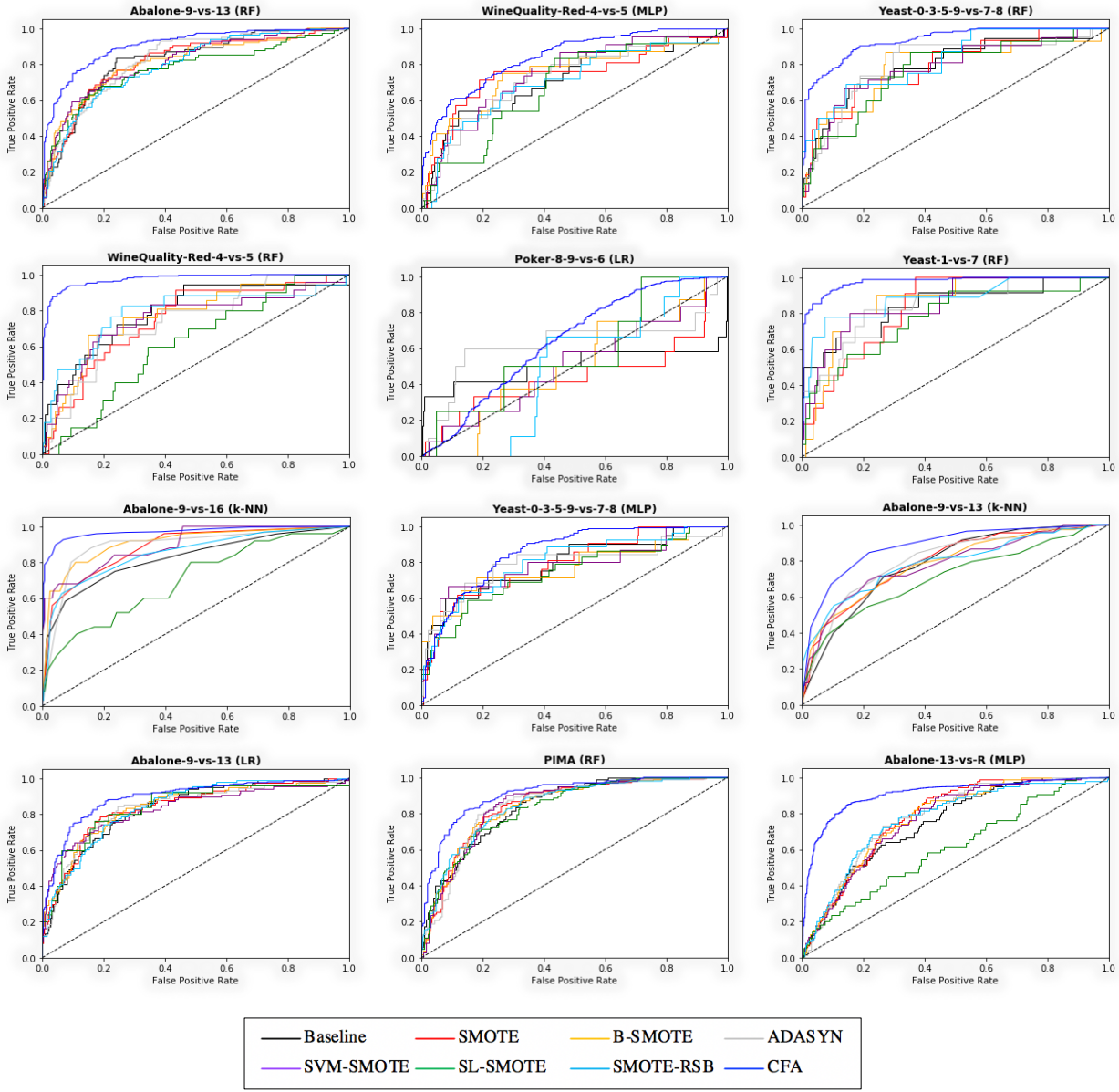

**Figure 3:** Selected examples for ROC curves where CFA outperformed SMOTE-based methods, obtained for the six methods using the four classifiers on different data sets.

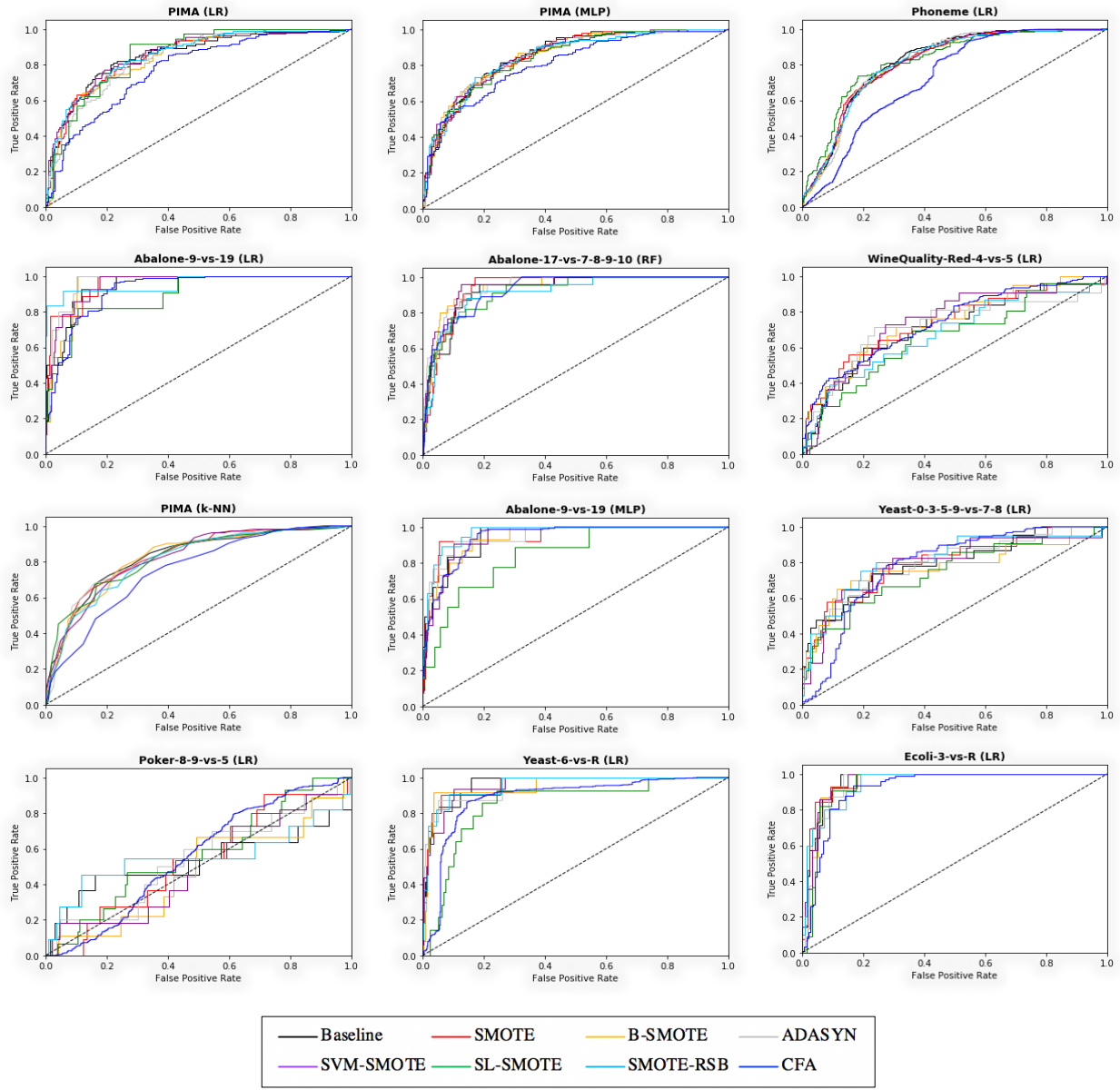

**Figure 4:** Selected examples for ROC curves where SMOTE variants outperformed CFA, obtained for the six methods using the four classifiers on different data sets.

**4.3.1 Why Does CFA Work?**

In XAI, counterfactual methods have been found to create plausible, synthetic datapoints for explanatory purposes; indeed, the evaluative metrics in XAI show that these explanatory counterfactuals are generally valid, within-distribution and close to existing data-points. This experience in XAI is the backdrop and motivation for applying this counterfactual method to the data augmentation. As we saw earlier, initial tests on a crop-growth prediction problem showed that generated counterfactuals in the minority class improved performance, specifically dealing with the dataset drift caused by climate change (see [67]). This experience led to the present tests to determine the generality of these effects. As we can see, the CFA method seems to work well across wide range of datasets, ML models and different imbalance ratios. But why does it work so well?

From the climate example, our initial explanation was that CFA does well because it generates minority instances that are "counterfactual offsets" from known minority-instances. But, this account does not answer why the offsets tend to be useful. Our best account hinges on ideas from how case-based reasoning (CBR) systems operate. In CBR, target problems are solved by retrieving similar cases and (sometimes) adapting them to generate predictions. So, if I am trying to predict house-prices in a city and my CBR system is presented with a "3-bed apartment with 4-bathrooms" and the closest retrieved case is a "3-bed apartment with 2-bathrooms", the system could have an adaptation rule than can bridges the gap between the historical case and the target case; for instance, there may be an adaptation rule that says "In general, an additional bathroom is worth $5k more". So, in a typical CBR-system, this rule would be applied to the retrieved case to bring it closer to the target case and improve the prediction. In CBR, adaptation rules were often hand-crafted but they may also be learned from analyses of feature-difference patterns between

instances in the case-base [15, 28, 75]; so, the extra-bathroom rule could be learned from bathroom-number differences found between historical instances (showing that they often lead to a $5k uplift in price, other features being equal). So, in CBR, these adaptation rules help to bridge holes in the case-base/dataset, by providing plausible transformations of known datapoints.

Counterfactuals are special case of an adaption rule; they capture the key feature-differences that lead to class changes across the boundary between majority and minority instances. So, when we apply them to majority instances to create synthetic minority instances, they stand a good chance of being plausible as they are based on prior local transformations. Though they lack generality (they are not created from multiple pairings of the same case-differences as is the case for learned adaption rules), perhaps they may work because they are so constrained and local. Remember, CFA only considers native counterfactual with $\leq 2$ feature differences, so the relationship is highly constrained and specific to the instances that are already very similar (i.e., all other features are essentially identical). To put it simply, CFA delivers good counterfactual rules that work locally to generate plausible datapoints that are predictively useful. So, it looks like the highly-constrained nature of the 2 feature-difference is important to the success of the method.

### 4.3.2 What Are CFA's Limitations?

The flip-side to the success question is the failure one: namely, when do we think CFA will fail and what limitations might it encounter. The current experiments a version of CFA that performs well, so it is not immediately clear what would lead CFA to fail. However, there are several conditions under which CFA is likely to be less good, with respect to (i) the quality of dataset differences, (ii) the use of the 2-feature-difference constraint and (iii) the use of different tolerances.

**Quality of the Dataset.** Fundamentally, CFA depends on the set of native counterfactuals in the dataset for its success. To put this another way, there needs to be a rich and diverse set of good counterfactual pairings between majority and minority instances either side of a clear decision boundary. Without these counterfactuals the ability to generate synthetic datapoints will be severely hampered. Current indications are that at least 5% of the majority class need to be involved in these native counterfactuals, in order to provide a basis for generating minority instances from the 95% of the majority class. However, we have not systematically tested how changes in this percentage affect performance. What we do know is that for many datasets the current parameters for CFA deliver good performance, so this factor might be quite robust to disruption.

**The 2-feature-difference Constraint**. A key hyperparameter for CFA is the constraint that the native counterfactuals build from the dataset involve no more than two feature-differences. This is a strong constraint that was made originally on psychological grounds in XAI [39]; that is, it has been shown that people prefer sparse counterfactuals with 2-3 feature differences, as they are easier to comprehend [22, 23]. However, this rationale from XAI does not apply to data augmentation. In data augmentation, the 2-feature-difference may work because it produces very minimally-different counterfactual pairs; so, they produce very simple adaptation rules in which most features remain the same between instances and a small number of features differ. These difference patterns may be more representative of valid instance-differences in the dataset and, hence, be more likely to produce good synthetic datapoints. These is some evidence to support this proposition in prior work. Temraz et al [67] report that in pilot runs of their experiments they explored using 3-, 4- and 5-feature-difference counterfactuals but found they did not significantly improve predictive importance; that is, they were less likely to generate useful minority instances. We do not know whether similar results would be found for other datasets, though the fact that the 2-feature-Temraz & Keane                                                Counterfactual Data Augmentationdifference constraint works here for 25 datasets with 9-12 features suggests that this constraint works quite generally. So, again, we would expect CFA to fail if higher numbers of feature-differences were used in computing the native counterfactuals when applying the method.

**The Importance of Tolerance**. A final key parameter in CFA is that of tolerance. In finding matching- and difference-features between two instances for a native counterfactual, we apply a tolerance to the feature values. Specifically, we allow features to match if their values are within +/-10% of the standard deviation from the mean all the values for that feature. This tolerance was applied uniformly across all of our datasets. Keane & Smyth [39] used a more sophisticated tolerance scheme that tailored the tolerance to each dataset; they varied the tolerances for each feature until changes in classification of the original dataset arose and then chose a relative tolerance that produced no classification change. Obviously, without tolerance, fewer counterfactuals would be found and the generative benefits of them would likely diminish.

**The Importance of the Decision Boundary.** Fundamentally, like Borderline-SMOTE, CFA works with instances that are close to the decision boundary. So, clearly, the definition and nature of that decision boundary is critical to its successful performance. If the instances around the boundary are noisy and the boundary is less clearly-defined then CFA is likely to disimprove. In this respect, it is interesting note that CFA does best using the *k*-NN and Random Forests models relative to the MLP and Linear Regression models, with the latter doing the worst (on AUC).

## 5  Conclusion

In this paper a novel oversampling method -- Counterfactual Augmentation (CFA) -- was proposed to handle the class imbalanced problem for binary classification tasks. CFA uses a case-based reasoning approach to generating synthetic counterfactuals in the minority class. The essence of this method is that it oversamples by adaptively combining actual feature-values from dataset instances rather than extrapolating/interpolating values between instances. The key discoveries made are: i) counterfactual methods developed for XAI can be usefully deployed to augment datasets, with synthetic cases in the minority class, that improve the performance of the ML models; (ii) this method can successfully introduce new synthetic minority examples by leveraging known counterfactuals in the dataset; and iii) this method can outperform many key benchmark SMOTE variants on a wide range of datasets with differing imbalance ratios using representative ML models.

## Acknowledgements

This publication has emanated from research conducted with the financial support of (i) Science Foundation Ireland (SFI) to the *Insight Centre for Data Analytics* under Grant Number 12/RC/2289_P2 and (ii) SFI and the Department of Agriculture, Food and Marine on behalf of the Government of Ireland to the *VistaMilk SFI Research Centre* under Grant Number 16/RC/3835.

Temraz & Keane                                    Counterfactual Data Augmentation